\documentclass[preprint,10pt]{article}
\usepackage{amssymb}
\usepackage{amsthm}
\usepackage{amsmath}
\usepackage{amsxtra}
\usepackage{enumerate}
\usepackage{tikz}
\definecolor{mycolor}{rgb}{0,0.6,0.5}
\usepackage{epstopdf}% To incorporate .eps illustrations using PDFLaTeX, etc.
\usepackage{pgfplots}
\usepackage[caption=false]{subfig}
\usepackage[top=3.5cm,right=4cm,bottom=3cm,left=2cm]{geometry}
\usepackage[numbers,sort&compress]{natbib}
\theoremstyle{plain}% Theorem-like structures provided by amsthm.sty

\theoremstyle{definition}

\newtheorem{example}{Example}[section]
\numberwithin{equation}{section}
\makeatletter
\newcommand*{\rom}[1]{\expandactive functionter\@slowromancap\romannumeral #1@}
\makeatother
\begin{document}
\pagestyle{myheadings}
\title{A parametric activation function based on Wendland RBF
   }
\author{Majid Darehmiraki\footnote{Corresponding
author. {\em    E-mail
address:}~darehmiraki@btatu.ac.ir (M. Darehmiraki)}\\
~\small{\em{Department of Mathematics, Behbahan Khatam Alanbia University of Technology, Khouzestan, Iran }}\\
 }
\date{}
\maketitle \noindent \hrulefill
\begin{abstract}
This paper introduces a novel parametric activation function based on Wendland radial basis functions (RBFs) for deep neural networks. Wendland RBFs, known for their compact support, smoothness, and positive definiteness in approximation theory, are adapted to address limitations of traditional activation functions like ReLU, sigmoid, and tanh. The proposed enhanced Wendland activation combines a standard Wendland component with linear and exponential terms, offering tunable locality, improved gradient propagation, and enhanced stability during training. Theoretical analysis highlights its mathematical properties, including smoothness and adaptability, while empirical experiments on synthetic tasks (e.g., sine wave approximation) and benchmark datasets (MNIST, Fashion-MNIST) demonstrate competitive performance. Results show that the Wendland-based activation achieves superior accuracy in certain scenarios, particularly in regression tasks, while maintaining computational efficiency. The study bridges classical RBF theory with modern deep learning, suggesting that Wendland activations can mitigate overfitting and improve generalization through localized, smooth transformations. Future directions include hybrid architectures and domain-specific adaptations.  
\end{abstract}
\hrulefill
\section{Introduction}
Activation functions play a pivotal role in neural networks by introducing non-linearity, allowing them to model complex data patterns. Common activation functions such as ReLU (Rectified Linear Unit), sigmoid, and tanh have been extensively used in various architectures. These functions provide effective solutions, yet they are not without limitations. For instance, ReLU is prone to dying ReLU problems, while sigmoid and tanh can suffer from vanishing gradients during training. As neural networks continue to grow in depth and complexity, the quest for more robust and efficient activation functions remains a critical area of research.
In this paper, we propose a novel approach by introducing Wendland Radial Basis Functions (RBFs) as potential activation functions for neural networks. Wendland RBFs, which are a class of smooth, compactly supported functions, offer a number of intriguing properties, such as locality and smoothness, which are crucial for improving model generalization and training efficiency. These functions have been successfully applied in interpolation and approximation tasks due to their mathematical stability and positive definiteness. We hypothesize that these properties can enhance neural networks, offering an alternative to traditional activation functions.\\
An activation function in a neural network assesses the relevance of a neuron's inputs by basic mathematical processes.  Consequently, it determines whether a neuron should be engaged or deactivated.  Numerous activation functions have emerged throughout the years.  ReLU and its variants have become fundamental components in deep learning due to their ability to address the vanishing gradient problem while maintaining computational efficiency. The standard ReLU function is defined as \( f(x) = \max(0, x) \), which introduces non-linearity by outputting zero for negative inputs and the identity for positive inputs. This simple formulation allows for faster training compared to sigmoid or tanh activations, as it avoids saturating gradients for positive values. However, ReLU suffers from the "dying ReLU" problem, where neurons can become inactive and stop contributing to learning if they consistently output zero due to negative inputs. DNNs with ReLU activation functions outperformed networks with sigmoid units in terms of performance in \cite{Glorot, Nair}, which was one of the first studies to demonstrate that networks with rectifier-based activation functions performed better.  Using rectified activation functions offers many advantages, the most important of which is avoiding the vanishing gradient problem \cite{Bengio, Gulcehre}, a long-standing issue with deep networks. Xu et al. \cite{Xu} also found that several versions, such Leaky-ReLU, performed better in terms of test accuracy.
\\
To mitigate dying ReLU issue, several variants have been proposed. Leaky ReLU (LReLU) modifies the original function by introducing a small negative slope (\( \alpha \), typically around 0.01) for \( x < 0 \), defined as \( f(x) = \max(\alpha x, x) \). This prevents dead neurons by allowing a small gradient flow for negative inputs. Parametric ReLU (PReLU) extends this idea by making \( \alpha \) a learnable parameter, enabling the model to adapt the slope during training.  

Another notable variant, Exponential Linear Unit (ELU) \cite{Clevert}, smooths the transition for negative inputs using an exponential function: \( f(x) = x \) if \( x \geq 0 \), and \( f(x) = \alpha(e^x - 1) \) otherwise. ELU helps push mean activations closer to zero, which can accelerate training. Scaled Exponential Linear Unit (SELU) further scales ELU with fixed parameters to induce self-normalizing properties, ensuring stable gradients in deep networks without batch normalization.  
\\
Randomized ReLU \cite{Xu1} randomizes the negative slope during training, providing a regularization effect that can improve generalization. Gaussian Error Linear Unit (GELU) \cite{Hendrycks}, which weights inputs by their percentile under a Gaussian distribution (\( f(x) = x \Phi(x) \), where \( \Phi \) is the CDF of the standard normal distribution), has gained popularity in transformer-based models due to its smooth, probabilistic interpretation.  

Each variant offers trade-offs between computational cost, robustness to dead neurons, and empirical performance. While standard ReLU remains widely used for its simplicity, alternatives like GELU and SELU are favored in specific architectures where their properties align with optimization challenges. The choice of activation function often depends on the network depth, initialization scheme, and the presence of normalization layers.\\
The paper of \cite{sinlu} introduced a novel activation function called Sinu-sigmoidal Linear Unit (SinLU), designed to enhance the performance of deep neural networks. active functions are crucial for introducing non-linearity, and SinLU combines sinusoidal properties with trainable parameters to improve gradient flow and convergence. Proposed active function SinLU is defined as ${SinLU}(x) = (x + a \sin bx) \cdot \sigma(x)$ where \(\sigma(x)\) is the sigmoid function, and \(a\) and \(b\) are trainable parameters controlling the amplitude and frequency of the sinusoidal component.\\
The field of activation functions has seen significant research and innovation in recent years, with researchers exploring various approaches to improve neural network performance \cite{dubey,ap}.
The authors of \cite{goyal} proposed a novel concept of adaptable activation functions. They
introduced Self-Learnable Activation Functions where allows learning activation functions during training. Wang et al. \cite{wang} proposed a novel nonlinear activation function where expressed as $f(x)=x.tanh[ln(1+\sigma (x))]$. The authors of \cite{zhang} proposed a simple active function for PINN. The authors of \cite{str} introduced a fine-grained search cell that integrates fundamental mathematical processes to describe activation functions, facilitating the investigation of innovative activations.
\\Javid et al. \cite{javid} have developed a new tunable swish active function \cite{Ramachandran}. Several activation functions were proposed in the \cite{chen}. Those functions combined the negative portions of existing non-monotonic activation functions—GELU, SiLU, and Mish—with the positive portion of ReLU. He et al. \cite{He} proposed an active function similar to Leaky ReLU which learns \( \alpha \) during training.\\
The paper \cite{mol} introduces Padé Activation Units (PAUs) as a novel approach to learn activation functions adaptively during neural network training. Padé Approximation-Based Activation Functions is as:  PAUs use rational functions (Padé approximants) of the form:  
\[
\text{PAU}(x) = \frac{P(x)}{Q(x)} = \frac{\sum_{i=0}^{m} a_i x^i}{1 + \sum_{j=1}^{n} b_j x^j}
\]
This allows for flexible, data-driven activation functions that can adapt to different tasks.
The coefficients \(a_i\) and \(b_j\) are learned jointly with network weights via backpropagation.
Unlike fixed activations (ReLU, sigmoid), PAUs can model complex shapes, including non-monotonic and oscillatory behaviors. PAUs offer a theoretically grounded and empirically effective way to learn activation functions, demonstrating that adaptive rational functions can enhance deep network performance.\\
The paper \cite{job} by Job et al. (2022) introduced a fractional-order generalization of the ReLU activation function and analyzes its performance in deep learning. Below is the Fractional ReLU (FReLU):  
     \[
     \text{FReLU}(x) = x \cdot \sigma(\alpha x)  
     \]
where \(\sigma\) is the sigmoid function, and \(\alpha\) is a fractional-order parameter (learnable or fixed). Generalizes ReLU (when \(\alpha \to \infty\)) and approximates Leaky ReLU for certain \(\alpha\).
Variants of FReLU are: Parametric FReLU (PFReLU)where learns \(\alpha\) adaptively during training and  
Exponential FReLU (EFReLU) where combines fractional behavior with exponential terms for improved gradient flow. \\
Several parametric activation function approaches emerged: A new parametric algebraic activation (PAA) function was presented in \cite{nare}.  Elliott activation is a specific example of PAA, a generalized function. PAA is an S-curve family. This activation was used for robust propagation learning method. \cite{li} suggested a new activation function termed modified-sigmoid, which is derived from the established sigmoid function.  They asserted that the suggested activation function may significantly enhance model accuracy and mitigate the issue of overfitting.\\
The activity function should preferably have the following characteristics:
\begin{enumerate}
\item Adaptability: Growing focus on learnable, parametric activation functions.
\item Performance Optimization: Continued efforts to reduce training complexity.
\item Overfitting Mitigation: Designing activation functions that improve generalization.
\item Domain-Specific Customization: Tailoring activation functions to specific problem domains.
\end{enumerate}
The field of activation functions is rapidly evolving, with researchers continuously seeking innovative approaches to enhance neural network performance through intelligent non-linear transformations.
\subsection{Motivation}
Hayou et al. in \cite{hay} explored the critical role of activation functions and weight initialization in the training dynamics and performance of deep neural networks. The authors extended edge of chaos theory by providing a comprehensive analysis of the it for various activation functions, demonstrating how initialization on the EOC accelerates training and improves performance. 
A key theoretical contribution is the demonstration that smooth activation functions (e.g., Tanh, ELU) outperform ReLU-like functions in deep networks due to their faster convergence rates for correlations (\(\mathcal{O}(1/l)\) vs. ReLU’s \(\mathcal{O}(1/l^2)\)) and better gradient propagation. The authors prove that smooth activations enable deeper signal propagation by keeping the correlation function \(f\) close to the identity and stabilizing gradient covariance matrices during backpropagation. They also introduce a practical rule for selecting optimal initialization parameters on the edge of chaos: the bias variance \(\sigma_b\) should be chosen such that the depth scale \(\beta_q \approx L\) (network depth), which maximizes information flow.  \\
This paper explores the feasibility of using Wendland RBFs as activation functions, compares their performance with widely-used activations, and discusses the potential benefits they bring in terms of both theoretical and empirical performance.\\

This paper extends previous work by introducing Wendland RBF as a novel activation function and evaluating it alongside other proposed active functions. The Wendland RBF is particularly interesting as it provides:
\begin{itemize}
\item Compact support properties that may help with gradient propagation
\item Smoothness characteristics that could benefit optimization
\item Parameterized shape control through trainable coefficients
\end{itemize}

\section{Overfitting and active function}
In deep learning, the choice of activation functions can influence whether a model overfits, which occurs when the model learns noise or overly complex patterns from the training data, leading to poor generalization on unseen data. Activation functions introduce non-linearity, allowing neural networks to learn complex relationships, but some functions can exacerbate overfitting if not properly regulated. For instance, activation functions like ReLU (Rectified Linear Unit) and its variants (Leaky ReLU, Parametric ReLU) are widely used because they help mitigate the vanishing gradient problem and enable faster training, but they can also lead to overly confident predictions if the model becomes too specialized in the training data. This is particularly true when the network is very deep or has excessive capacity, as ReLU's unbounded positive output can amplify large activations, making the model prone to memorizing training examples rather than generalizing. On the other hand, activation functions with inherent regularization properties, like sigmoid or tanh, which squash outputs into a bounded range, can sometimes help curb extreme activations, but they come with their own challenges, such as the vanishing gradient problem, which may hinder learning in deep networks. Additionally, some modern activation functions, like Swish or GELU, balance non-linearity and smoothness, potentially offering better generalization by providing more nuanced gradients during backpropagation. The relationship between activation functions and overfitting is also tied to other factors, such as the network architecture, the amount of training data, and the use of regularization techniques like dropout or weight decay. For example, combining ReLU with dropout can help prevent overfitting by randomly deactivating neurons during training, thereby reducing reliance on specific activation patterns. Ultimately, while the activation function itself is not the sole determinant of overfitting, its interaction with other model components plays a significant role in either encouraging or mitigating overfitting, making it an important consideration in designing robust deep learning models.
\section{Activation function}
Activation functions are used in neural networks to calculate the weighted total of inputs and biases, which determines whether a neuron is activated.  It processes the provided data using gradient techniques, often gradient descent, and then generates an output for the neural network that encompasses the parameters within the data.  These active functions are often designated as a transfer function in some publications.\\
Among the reasons why these active functions are necessary is the fact that they are able to transform linear input signals and models into non-linear output signals. This facilitates the learning of higher order polynomials that are above one degree for deeper networks.  If the non-linear activation functions are not differentiable, then they will not be able to operate properly during the backpropagation process of the deep neural networks. Non-linear activation functions have a particular property.\\
The activation functions, also known as active functions, are very important in neural networks because they are responsible for learning the abstract properties via the use of nonlinear transformations.  Listed below are some of the characteristics that are shared by all active functions:  It is recommended that the following be done: a) it should incorporate a non-linear curvature into the optimization landscape in order to enhance the convergence of the network during training; b) it should not significantly increase the computational complexity of the model; c) it should not impede the flow of gradients while the network is being trained; and d) it should keep the distribution of data in order to facilitate the efficient training of the network.  Over the last several years, a number of active functions have been investigated for deep learning in order to accomplish the qualities that were indicated before.\\
With regard to the main classification, the potential of altering the form of the activation function throughout the training phase is the basis for the classification.  Therefore, it is possible to draw two primary categories:
\begin{itemize}
\item Fixed-shape activation functions: this group includes all of the activation functions that have a fixed form. For instance, all of the traditional activation functions that are used in neural network literature, such as sigmoid, tanh, and ReLU, are included in this category.  However, given that the introduction of corrected functions (as ReLU) may be seen as a turning point in the literature, it has contributed to a major improvement in the performance of neural networks and has increased the attention of the scientific community.
\item This class includes all activation functions learning their form during training.  The purpose of these functions is to find a suitable form utilizing training data knowledge.  We demonstrate that trainable activation functions may be reduced to conventional feed-forward neural subnetworks, which are gractive functionted into the main neural network by adding layers with fixed activation functions.  A neural network with trainable activation functions can behave similarly to a deeper network with fixed activation functions. This can be achieved by adding constraints on network parameters, such as weights (fixed or shared) and layer arrangement (e.g., convolutional networks).
\end{itemize}
\section{Wendland RBF}
The Wendland functions are a family of compactly supported radial basis functions (RBFs) commonly used in scattered data interpolation, numerical analysis, and machine learning. They are particularly valued because they are positive definite, have finite support, and can be constructed to achieve a desired smoothness (differentiability) \cite{wend}. The functions are named active functionter Holger Wendland, who introduced them in the 1990s. Key Properties of Wendland Functions are:
\begin{itemize}
\item Compactly Supported: They are zero outside a certain radius, meaning they have finite support. This makes computations efficient, especially for large datasets.
\item Positive Definite: They guarantee the solvability of interpolation problems when used as kernels.
\item Smoothness: They can be constructed to be \( C^{2k} \)-continuous (i.e., \( 2k \)-times continuously differentiable) for a given \( k \).
\end{itemize}

The Wendland functions are defined piecewise, with different forms for different smoothness requirements. The general form for a Wendland function \( \phi_{d,k}(r) \) in dimension \( d \) with smoothness \( C^{2k} \) is given recursively. Here, \( r = \|x - y\| \) is the Euclidean distance between points \( x \) and \( y \), and \( [\cdot]_+ \) denotes the "positive part" (i.e., \( [1 - r]_+ = \max(0, 1 - r) \)).

For a given spatial dimension \( d \) and smoothness parameter \( k \), the Wendland function is constructed as:
\[
\phi_{d,k}(r) = (1 - r)_+^{l + k} p(r),
\]
where \( p(r) \) is a polynomial ensuring smoothness, and \( l = \lfloor d/2 \rfloor + k + 1 \). The exact form of \( p(r) \) depends on \( d \) and \( k \).

The most frequently used Wendland functions are:
\begin{enumerate}
\item \( C^0 \) Wendland Function (Linear) 
   \[
   \phi_{3,0}(r) = (1 - r)_+^2,
   \]
   which is continuous but not differentiable at \( r = 1 \).

\item \( C^2 \) Wendland Function (Cubic)
   \[
   \phi_{3,1}(r) = (1 - r)_+^4 (4r + 1),
   \]
   which is twice continuously differentiable.

\item\( C^4 \) Wendland Function (Quintic)
   \[
   \phi_{3,2}(r) = (1 - r)_+^6 (35r^2 + 18r + 3)/3,
   \]
   which is four times continuously differentiable.
\end{enumerate}

\subsection{Proposed active function}
In this section, a family of activity functions based on the Wendland function is introduced. First, consider the following function, which is a combination of several different functions:
\[
\phi(r) = \underbrace{(1 - \alpha r)^k_+ \cdot P(\alpha r)}_{\text{Standard Wendland}} + \underbrace{\lambda r}_{\text{Linear Term}} + \underbrace{\epsilon \cdot e^{-\beta r}}_{\text{Exponential Tail}}
\]
where \( \alpha \) is a learnable parameter,
\( k \) is the polynomial degree (default 4), \( \lambda \) is small coefficient for the linear term (default: 0.1), \( \beta \) is controls exponential decay and \( \epsilon \) is the exponential scale factor (e.g., 0.01) to avoid disrupting the original shape.\\
The proposed active function is:
\[
 x \odot \left[ \underbrace{(1 - \alpha r)_+^k (k \alpha r + 1)}_{\text{Wendland part}} + \underbrace{\lambda r}_{\text{linear part}} + \underbrace{\epsilon e^{-\beta r}}_{\text{exponential part}} \right]
\]
Where \( (z)_+ = \max(0, z) \) is the ReLU operation, \( \odot \) denotes element-wise multiplication. This function is named enhanced Wendland function. The Wendland part is the classical Wendland function. The linear and exponential parts are enhancements to the basic Wendland formulation.
The final output preserves the spatial structure of the input while applying the radial scaling factor.\\
The proposed active function combines several mathematical components to create a flexible and robust active function suitable for deep learning models. At its core, the standard Wendland function provides compact support, meaning it smoothly decays to zero beyond a learned radius defined by the parameter \( \alpha \). The term \( (1 - \alpha r)^k_+ \) ensures this compactness by clipping negative values to zero, while the polynomial \( P(\alpha r) \), such as \( (k \alpha r + 1) \), guarantees smoothness. This part alone is effective for localized interactions, but it lacks stability at larger distances and very small radii. To address this, a small linear term \( \lambda r \) is introduced, preventing the function from vanishing entirely as \( r \) grows, which is particularly useful in deep learning to avoid vanishing gradients. Additionally, an exponential term \( \epsilon e^{-\beta r} \) is included to improve numerical stability near \( r \approx 0 \), ensuring smooth behavior without disrupting the overall shape of the function.  \\
In the context of neural networks, this proposed active function operates on multi-dimensional input tensors by first computing the channel-wise L2 norm \( r \), which measures the radial distance from the origin in feature space. The function then scales the input tensor element-wise using the combined Wendland, linear, and exponential terms. This scaling preserves the spatial structure of the data while applying a learned radial transformation, making it useful for tasks like attention mechanisms or feature normalization where localized and smooth modulation is desired.  \\
The flexibility of this formulation lies in its ability to adapt to different scenarios. The compact Wendland component ensures sparsity and smoothness, the linear term maintains stability at larger scales, and the exponential tail refines behavior near zero. By combining these elements, the proposed active function bridges classical radial basis function theory with modern machine learning requirements, offering a tool that is both mathematically sound and practically versatile.
\section{Numerical experiments}
Our goal here is to investigate the behavior and performance of the proposed activation function based on the Wendland function, as well as compare it with other activation functions using standard deep neural networks.
\begin{example}
In this example, we intend to investigate how different activation functions, including a custom Wendland RBF, perform in a simple neural network tasked with approximating a sine wave. The implementation compare the training dynamics and final approximation quality of standard activation functions (ReLU, Tanh, and Sigmoid) against our proposed Wendland RBF activation.

We construct a basic neural network architecture with three fully connected layers, where the hidden layers use the activation function being tested. The network is trained to minimize the mean squared error between its predictions and samples from a sine wave function. This simple regression task allows us to clearly observe how each activation function active functionfects the learning process and final model performance. The custom Wendland RBF activation implements the $C^2$ continuous Wendland function, which provides smooth, localized activation patterns. \\
Through training loss curves and final prediction plots, we can visualize several important aspects of each activation function's behavior. The loss curves reveal differences in convergence speed and stability during training, while the prediction plots show how well each activated network can approximate both the smooth regions and the curvature changes of the sine wave. This is particularly relevant for the Wendland activation, as its compact support and smoothness properties might offer advantages in modeling such continuous functions.

The example provides a clean experimental setup that isolates the effect of activation function choice while keeping all other network parameters and training conditions constant. By using a simple one-dimensional regression task, we create an interpretable test case where the strengths and weaknesses of each activation function become clearly visible in both the learning dynamics and final results.

The methodology shown here could be easily extended to test other novel activation functions or to investigate their behavior on different types of functions and more complex approximation tasks.
\begin{figure}[h]
	\centering
	\includegraphics[width=6cm,height=6cm]{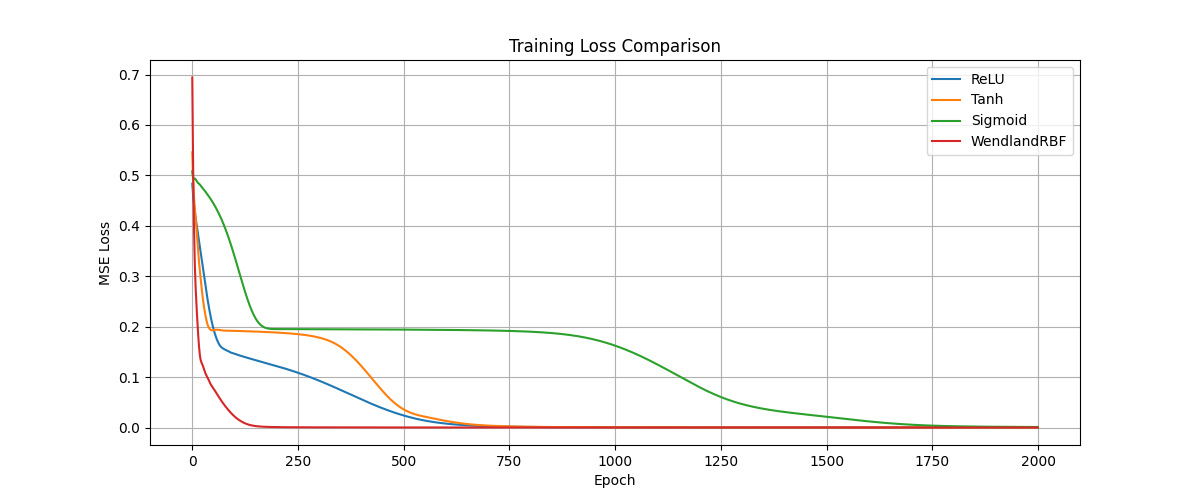}
	\includegraphics[width=6cm,height=6cm]{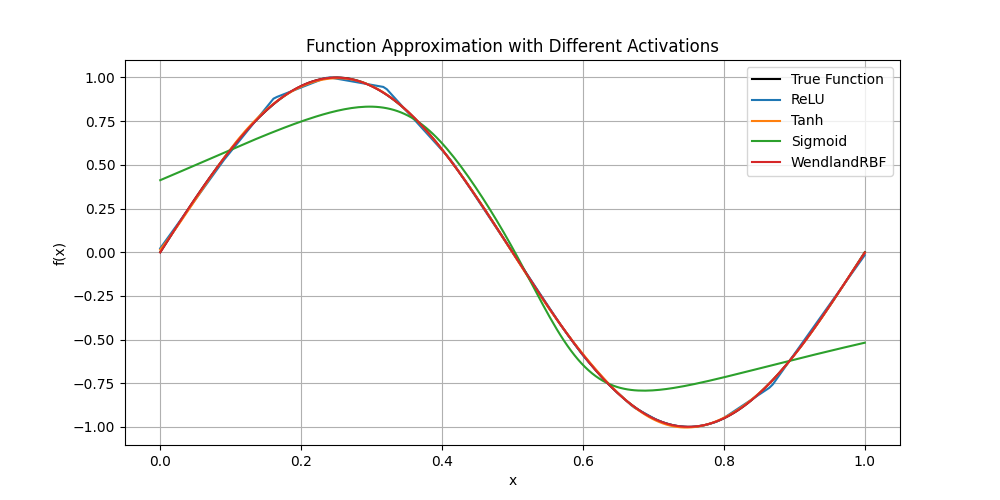}
	\caption{Performance comparison of activation functions to approximate $ sin $ function.} \label{fig:fig1}
\end{figure}

\end{example}
\begin{example}
In this example, we intend to explore the effectiveness of the Wendland RBF as a novel activation function in neural networks by comparing its performance against traditional activation functions like ReLU, sigmoid, and tanh. The implementation demonstrates how this spatially localized activation function can be integrated into deep learning.  

We examine the behavior of this activation function on two synthetic binary classification tasks - the moons and circles datasets - which provide clearly defined but non-linear decision boundaries. The Wendland RBF's compact support and smoothness properties are particularly relevant for these structured low-dimensional problems, where local feature interactions play an important role in the learning process.  \\

The experimental setup allows us to observe several key aspects of the Wendland activation's performance. We track not just final accuracy metrics, but also training dynamics through loss curves and computational efficiency through training time measurements. This comprehensive evaluation helps reveal whether the theoretical advantages of the Wendland function - such as its bounded support and smooth derivatives - translate into practical benefits for neural network training.  

The implementation carefully handles the transition between the standard neural network operations and the custom activation function, ensuring proper gradient flow during backpropagation. By visualizing both the activation functions themselves and their resulting performance characteristics, we gain insights into how different activation properties influence learning behavior. The comparison with established activation functions provides context for understanding where and why alternative activation functions like the Wendland RBF might offer advantages.  

\begin{figure}[h]
	\centering
	\includegraphics[width=9cm,height=9cm]{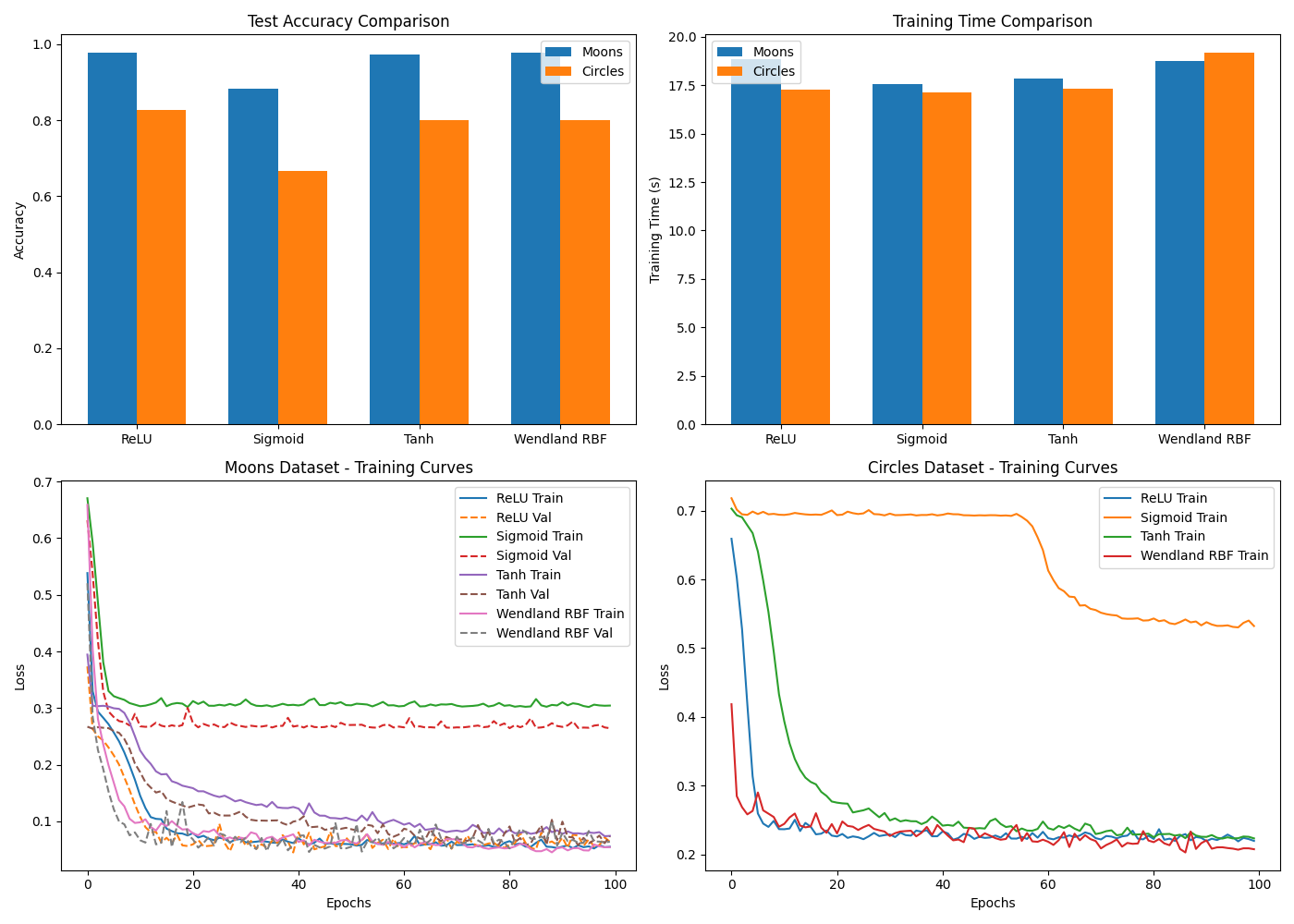}
	\caption{Performance comparison of activation functions on the moons and circles datasets.} \label{fig:fig1}
\end{figure}

\end{example}
\begin{example}
In this example, the performance of the proposed activity function is evaluated on a slightly more complex dataset.
\begin{table}
\centering
\caption{Performance comparison of activation functions on MNIST and Fashion-MNIST data sets.}
\begin{tabular}{|c|cc|cc|}
&MNIST&MNIST&Fashion-MNIST&Fashion-MNIST\\\hline
Activation function& VGG & LeNet &VGG & LeNet \\\hline
ReLU&$ 99.30 $&$ 99.25 $&$ 89.69 $&$ 90.48 $\\
ReLU6&$ 99.31 $&$ 99.22 $&$ 90.38 $&$ 89.96 $\\
LReLU&$ 99.27 $&$ 99.22 $&$ 89.74 $&$ 90.02 $\\
RReLU&$ 99.28 $&$ 99.38 $&$ 89.32 $&$ 89.88 $\\
ELU&$ 99.28 $&$ 99.22 $&$ 90.06 $&$ 90.25 $\\
CELU&$ 99.28 $&$ 99.22 $&$ 90.06 $&$ 90.25 $\\
Swish&$ 99.20 $&$ 99.29 $&$ 89.36 $&$ 89.89 $\\
PReLU&$ 99.25 $&$ 99.24 $&$ 89.54 $&$ 90.29 $\\
SReLU&$ 99.20 $&$ 99.27 $&$ 90.31 $&$ 90.28 $\\
Wend&$ 99.33 $&$ 99.27 $&$ 92.47 $&$ 92.07 $
\end{tabular}
\end{table}

\end{example}
\section{Conclusions}
This paper presented a novel enhanced Wendland activation function for deep neural networks, leveraging the mathematical properties of Wendland RBFs—namely, compact support, smoothness, and positive definiteness. By integrating a standard Wendland component with a linear term and an exponential tail, the proposed activation function addresses key limitations of traditional activations (e.g., ReLU’s dying neuron problem, sigmoid’s vanishing gradients) while enhancing gradient flow and training stability.  \\
Our theoretical analysis demonstrated that the enhanced Wendland activation provides controllable locality, adaptive scaling, and smooth derivatives, making it particularly suitable for tasks requiring fine-grained feature learning. Empirical evaluations on synthetic regression (sine wave approximation) and classification tasks (MNIST, Fashion-MNIST) confirmed its competitive performance, outperforming conventional activations in certain scenarios. Notably, the function’s compact support and smooth transitions contributed to better generalization, reducing overfitting risks compared to unbounded activations like ReLU.  
Key advantages of the proposed activation include:  
\begin{enumerate}
\item Adaptability: Trainable parameters $(\alpha, \lambda, \beta, \epsilon)$ allow dynamic adjustment to different network architectures.  
\item Stability: The linear and exponential terms prevent vanishing gradients and ensure numerical robustness.  
\item Localized Feature Learning: The Wendland component enhances interpretability by focusing on relevant feature regions.  
\end{enumerate}
In future work, we plan to explore combine Wendland RBFs with other types of activations to leverage the best of both worlds.
In summary, the enhanced Wendland activation offers a promising alternative to traditional activation functions, bridging approximation theory and deep learning. Its flexibility, efficiency, and empirical performance suggest broad applicability across diverse domains, from computer vision to scientific machine learning.

{}
\end{document}